\documentclass{article} 

\usepackage[dvipsnames]{xcolor}
\usepackage{iclr2025_conference,times} %

\usepackage{hyperref}
\usepackage{url}
\usepackage{tcolorbox}
\usepackage{natbib}
\usepackage[utf8]{inputenc} 
\usepackage[T1]{fontenc}    
\usepackage{booktabs}       
\usepackage{amsfonts}       
\usepackage{nicefrac}       
\usepackage{microtype}      

\usepackage{amssymb}
\usepackage{amsmath}
\usepackage{amsthm}
\usepackage{graphicx}
\usepackage{textcomp}
\usepackage{balance}
\usepackage{multirow}
\usepackage{subfigure}
\usepackage{array}
\usepackage{enumitem}
\usepackage{epstopdf}
\usepackage{threeparttable}
\usepackage{bm}
\usepackage{tabularx}
\usepackage{subcaption}

\newtheoremstyle{mydefn}
{}{}
{\it}       
{0pt}       
{\bfseries} 
{:~}        
{0pt}       
{}          
\theoremstyle{mydefn}

\newtheoremstyle{myexample}
{}{}
{}          
{0pt}       
{\bfseries} 
{:~}        
{0pt}       
{}          
\theoremstyle{myexample}

\newtheorem{example}{Example}

\renewcommand{\paragraph}[1]{\noindent\textbf{#1.}}
\renewcommand{\subparagraph}[1]{\noindent\textbf{\underline{#1.}}}

\newcommand{\ip}[1]{\langle #1 \rangle}

\def\T#1{{``\textcolor{Blue}{\textit{#1}}''}} 
\def\TwoQ#1{{\textcolor{Blue}{\textit{#1}}}}  
\def\Q#1{{``\textcolor{BrickRed}{\textit{#1}}''}}      
\def\yes{{``\textcolor{YellowOrange}{\textit{yes}}''}} 
\def\no{{``\textcolor{OliveGreen}{\textit{no}}''}}     

\def\algo#1{{\textbf{#1}}} 
\def\chkp#1{{\texttt{#1}}} 
\def\data#1{{\textbf{#1}}} 

\makeatletter
\newif\if@restonecol
\makeatother

\usepackage[linesnumbered,ruled,vlined]{algorithm2e}
\usepackage{algpseudocode}

\SetKwComment{Comment}{$\triangleright$\ }{}
\SetKwRepeat{Do}{do}{while}

\makeatletter
\def\UrlAlphabet{%
    \do\a\do\b\do\c\do\d\do\e\do\f\do\g\do\h\do\i\do\j%
    \do\k\do\l\do\m\do\n\do\o\do\p\do\q\do\r\do\s\do\t%
    \do\u\do\v\do\w\do\x\do\y\do\z\do\A\do\B\do\C\do\D%
    \do\E\do\F\do\G\do\H\do\I\do\J\do\K\do\L\do\M\do\N%
    \do\O\do\P\do\Q\do\R\do\S\do\T\do\U\do\V\do\W\do\X%
    \do\Y\do\Z}
\def\UrlDigits{\do\1\do\2\do\3\do\4\do\5\do\6\do\7\do\8\do\9\do\0}
\g@addto@macro{\UrlBreaks}{\UrlOrds}
\g@addto@macro{\UrlBreaks}{\UrlAlphabet}
\g@addto@macro{\UrlBreaks}{\UrlDigits}
\makeatother

\title{A General Framework for Producing Interpretable Semantic Text Embeddings}

\author{Yiqun Sun, Qiang Huang, Yixuan Tang, Anthony K. H. Tung \\
National University of Singapore\\
Singapore \\
\texttt{\{sunyq, huangq, yixuan, atung\}@comp.nus.edu.sg} \\
\And
Jun Yu \\
Harbin Institute of Technology (Shenzhen) \\
Shenzhen, China \\
\texttt{yujun@hit.edu.cn} \\
}

\iclrfinalcopy 
\begin{document}

\maketitle

\begin{abstract}
Semantic text embedding is essential to many tasks in Natural Language Processing (NLP).
While black-box models are capable of generating high-quality embeddings, their lack of interpretability limits their use in tasks that demand transparency.
Recent approaches have improved interpretability by leveraging domain-expert-crafted or LLM-generated questions, but these methods rely heavily on expert input or well-prompt design, which restricts their generalizability and ability to generate discriminative questions across a wide range of tasks.
To address these challenges, we introduce \algo{CQG-MBQA} (Contrastive Question Generation - Multi-task Binary Question Answering), a general framework for producing interpretable semantic text embeddings across diverse tasks.
Our framework systematically generates highly discriminative, low cognitive load yes/no questions through the \algo{CQG} method and answers them efficiently with the \algo{MBQA} model, resulting in interpretable embeddings in a cost-effective manner.
We validate the effectiveness and interpretability of \algo{CQG-MBQA} through extensive experiments and ablation studies, demonstrating that it delivers embedding quality comparable to many advanced black-box models while maintaining inherently interpretability. Additionally, \algo{CQG-MBQA} outperforms other interpretable text embedding methods across various downstream tasks.
\end{abstract}


\section{Introduction}
\label{sect:intro}

Text embedding is a cornerstone of Natural Language Processing (NLP), transforming texts—whether sentences, paragraphs, or full documents—into embedding vectors that capture their semantic meaning.
In semantic embedding spaces, the similarity between texts is represented by the proximity of their embedding vectors, typically measured using distance measures like Euclidean distance, cosine distance, or inner product. 
The closer the vectors, the more semantically similar the texts. 
These embeddings are foundational to many downstream NLP tasks, including Semantic Textual Similarity (STS) \citep{agirre-etal-2012-semeval, agirre-etal-2013-sem}, Information Retrieval \citep{karpukhin-etal-2020-dense, thakur2beir}, Clustering \citep{aggarwal2012survey}, and more recently, Retrieval Augmented Generation (RAG) \citep{lewis2020retrieval, guu2020realm, asai2024self}. 

Black-box text embedding methods, such as \algo{Sentence-BERT} \citep{reimers-gurevych-2019-sentence}, \algo{SimCSE} \citep{gao-etal-2021-simcse}, \algo{WhitenedCSE} \citep{zhuo-etal-2023-whitenedcse}, and \algo{AnglE} \citep{li-li-2024-aoe}, excel at generating high-quality embeddings by training on vast amounts of data. 
These models are highly effective at capturing semantic similarities, making them indispensable for a variety of NLP tasks \citep{muennighoff-etal-2023-mteb}.
However, their black-box nature leaves the embeddings opaque to human users.
These models do not provide insight into why certain texts are deemed similar, which becomes problematic for tasks that require transparency, especially in applications involving high-stakes decision-making, such as legal and medical domains, or in cases requiring explanations for regulatory compliance.

Interpretability in machine learning is the ability of humans to understand the reasoning behind a model's results \citep{MILLER20191}, which is essential not only for building trust and ensuring safety but also for detecting biases and debugging models \citep{molnar2022}. 
Recent advances have enhanced interpretability by leveraging inherently interpretable models such as \algo{Decision Tree} \citep{breiman1984classification} and \algo{Generalized Additive Models} \citep{GAM}, as well as model-agnostic methods like \algo{LIME} \citep{lime} and \algo{SHAP} \citep{shap}. 
However, these interpretable approaches lose effectiveness when applied on top of non-interpretable features generated by black-box text embedding models.
Consequently, the challenge remains: how can we create interpretable text embeddings without sacrificing performance?

Recent efforts have sought to address the challenge of creating interpretable embeddings by using questions as interpretable dimensions.
For instance, \algo{ChiLL} \citep{mcinerney-etal-2023-chill} employs yes/no questions crafted by domain experts to classify patient clinical notes, but its reliance on costly expert annotation limits its generalizability to different datasets. 
\algo{QAEmb} \citep{Benara2024CraftingIE} advances this concept by using task-specific prompts with examples to automatically generate yes/no questions via Large Language Models (LLMs), achieving notable success in the fMRI prediction task \citep{huth2016natural, lebel2023natural, tang2023semantic}. 

Nonetheless, \algo{QAEmb} requires meticulously crafted prompts and uses six distinct prompt templates to generate questions for the fMRI prediction task, which complicates its usage in general settings due to the need for prompt engineering expertise. 
Furthermore, this example-based question generation approach often produces generic, less discriminative questions, limiting its effectiveness in broader applications.
Given the importance of question quality in creating effective interpretable embeddings, there is a pressing need for a systematic approach that can automatically generate meaningful and discriminative questions across various text embedding tasks.

To address this gap, we introduce \algo{CQG-MBQA} (Contrastive Question Generation - Multi-task Binary Question Answering), a \emph{general} framework for producing interpretable semantic text embedding, which matches the performance of many black-box models and surpasses existing interpretable baselines across various text embedding tasks.
\algo{CQG-MBQA} harnesses contrastive learning principles to prompt LLMs to generate highly discriminative binary yes/no questions, which form the dimensions of the embedding space. 
These questions not only capture the semantic nuances between texts but also offer human-interpretable insights.
The main contributions of this work are as follows:
\vspace{-0.25em}
\begin{itemize}[nolistsep,leftmargin=25pt]
  \item We propose \algo{CQG-MBQA}, the first general framework that tackles the challenge of generating interpretable text embeddings for a broad range of tasks, offering a practical and scalable solution for text representation.

  \item Our \algo{Contrastive Question Generation (CQG)} method produces highly discriminative questions that offer high interpretability while minimizing cognitive load for users, ensuring that the semantic relationships between texts can be easily understood.
 
  \item The \algo{Multi-task Binary Question Answering (MBQA)} model processes these binary questions efficiently at scale, significantly reducing the inference costs typically associated with LLMs, making the framework cost-effective for real-world applications.

  \item We validate the effectiveness of \algo{CQG-MBQA} through extensive experiments and ablation studies, demonstrating its robustness and practical applicability across multiple benchmarks and downstream tasks.
\end{itemize}
\vspace{-0.3em}
\section{Related Work}
\label{sect:related}

\vspace{-0.3em}
Text embedding is a fundamental NLP task that transforms texts into vector representations that capture their semantic meanings. 
Generally, it can be classified into two types: black-box embedding and interpretable embedding.

\paragraph{Black-box Embedding}
Early methods for text embedding, such as \algo{GloVe} \citep{pennington-etal-2014-glove} and \algo{Word2Vec} \citep{NIPS2013_9aa42b31}, typically pool word embeddings to create low-dimensional semantic representations. 
However, these methods, which rely on individual word embeddings, often fail to capture the full context of a text. 
For example, the sentences \T{Most people in the world \textbf{like} Apple.} and \T{Most people in the world \textbf{do not like} Apple.} share high lexical overlap but have opposite meanings, highlighting the limitations of such methods, which struggle to capture deeper semantic differences beyond surface-level word similarity.

To produce context-aware text embeddings, \algo{Universal Sentence Encoder (USE)} \citep{cer-etal-2018-universal} employs a transformer model \citep{transformer} trained on a combination of unsupervised tasks and supervised fine-tuning using the Stanford Natural Language Inference (SNLI) corpus \citep{snli}. 
\algo{BERT} \citep{devlin-etal-2019-bert}, a transformer network pre-trained on large-scale unlabeled text, can generate sentence embeddings by pooling its output representations.
Subsequent models have further refined \algo{BERT} using contrastive learning and other semantic-related objectives. 
For instance, \algo{Sentence-BERT (SBERT)} \citep{reimers-gurevych-2019-sentence} pioneers the Siamese network structure for Semantic Textual Similarity (STS), while \algo{SimCSE} \citep{gao-etal-2021-simcse} develop a contrastive learning framework for both unsupervised and supervised settings. 
\algo{WhitenedCSE} \citep{zhuo-etal-2023-whitenedcse} enhances embedding uniformity and alignment with shuffled group whitening, and \algo{AnglE} \citep{li-li-2024-aoe} optimizes angle differences to overcome cosine similarity limitations. 
Despite these advances, black-box models produce embeddings that are opaque and difficult to interpret. In this work, we target generating interpretable dimensions for text embedding.

\paragraph{Interpretable Embedding}
The challenge of creating interpretable embeddings has been persisted, especially with the rise of dense word embeddings. 
Early efforts focus on transforming word embeddings to improve interpretability. 
\cite{jha2018interpretable} apply categorical knowledge in the biomedical domain to convert pre-trained embeddings into interpretable dimensions, while \cite{senel2018semantic} quantify interpretability by analyzing latent semantic structures. 
Models like \algo{SPINE} \citep{subramanian2018spine} employ auto-encoders to create interpretable embeddings from non-interpretable ones like \algo{GloVe} \citep{pennington-etal-2014-glove} and \algo{Word2Vec} \citep{NIPS2013_9aa42b31}, and \algo{Word2Sense} \citep{panigrahi-etal-2019-word2sense} creates interpretable dimensions based on specific word senses.

Despite progress, developing context-aware, interpretable dimensions remains difficult. 
Recent research has shifted towards indirectly understanding embedding spaces. 
For instance, \cite{lee-etal-2022-toward} introduce token pair contribution heatmaps to enhance interpretability in sentence similarity, while \cite{simhi-markovitch-2023-interpreting} propose transforming embedding spaces into comprehensible conceptual representations. 
Recent advancements like \algo{ChiLL} \citep{mcinerney-etal-2023-chill} generate interpretable binary features from health records by querying pre-trained LLMs with \emph{expert-crafted} yes/no questions for patient classification. 
\algo{QAEmb} \citep{Benara2024CraftingIE} extends this by prompting LLMs to \emph{automatically} generate questions using examples of texts and questions, demonstrating its efficacy in the fMRI prediction task. 
Inspired by \algo{QAEmb}, we propose a cost-effective framework that generates high-quality questions and uses them as interpretable dimensions for text embedding. 

\section{Interpretable Text Embedding Framework}
\label{sect:framework}

We introduce an interpretable text embedding framework named \textbf{CQG-MBQA}, which uses questions as interpretable dimensions to represent text. 
By posing a set of carefully designed yes/no questions about a given text, the answers form an interpretable embedding vector that captures the text's core semantics. 
For instance, consider three questions: \Q{Is the article about AI?}, \Q{Is the article about sports?}, and \Q{Is the article about food?}. 
For the text \T{Apple is a technology company.}, querying an LLM yields the answers: [\yes, \no, \no], resulting in the embedding vector $[1, 0, 0]$, which reflects the text's key features. 
Applying the same set of questions across all texts in a corpus produces a consistent embedding matrix that encodes the semantic information of the entire dataset.

\begin{figure}[t]
  \centering
  \includegraphics[width=0.99\textwidth]{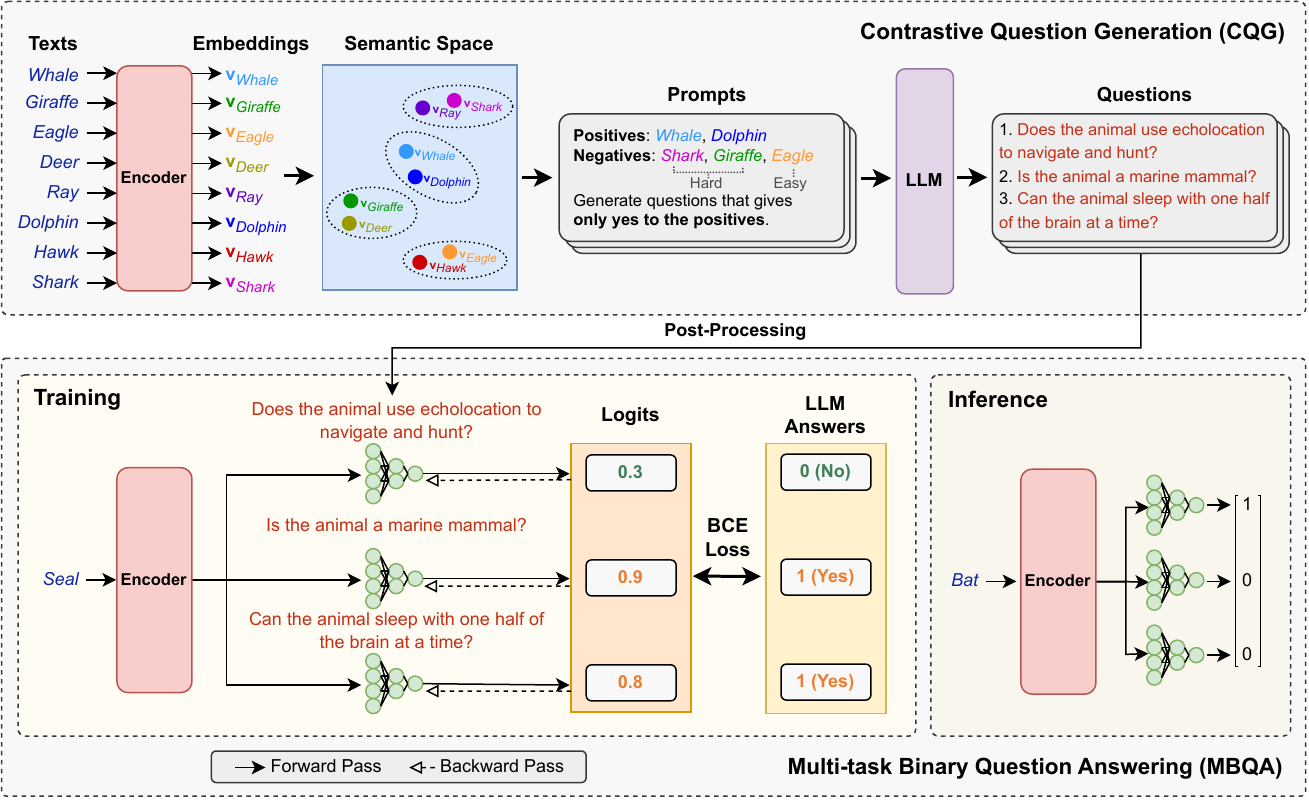}
  \vspace{-0.5em}
  \caption{An overview of the \algo{CQG-MBQA} framework.}
  \label{fig:framework}
  \vspace{-1.0em}
\end{figure}

As shown in Figure \ref{fig:framework}, the \textbf{CQG-MBQA} framework consists of two phases: question generation and question answering. 
To generate high-quality, discriminative questions, we develop a method called \textbf{Contrastive Question Generation (CQG)}, which harnesses pre-trained dense text embedding models and generative LLMs for question generation. Details of this method are outlined in Section \ref{sect:framework:generation}.
Once the questions are generated, their corresponding answers form the text's embedding vector. 
Yet, generating answers through LLMs at scale is both time-consuming and expensive. To address this, we propose a \textbf{Multi-task Binary Question Answering (MBQA)} model. Trained with a multi-task binary classification objective, this model can generate interpretable embeddings efficiently, requiring far fewer LLM API calls. 
Further details on this model are provided in Section \ref{sect:framework:answer}.

\subsection{Question Generation}
\label{sect:framework:generation}

\paragraph{Motivations}
To effectively represent texts using binary question answers, we need highly discriminative questions that capture the subtle semantic differences between texts in the corpus. 
Existing methods, such as \algo{QAEmb} \citep{Benara2024CraftingIE}, generate questions by prompting LLMs with dataset descriptions, example texts, and sample questions. 
However, this example-based approach presents two significant limitations:
\vspace{-0.25em}
\begin{enumerate}[nolistsep,leftmargin=30pt,label*=(\arabic*)] 
  \item \textbf{Insufficient Specificity:} 
  The generated questions are often too generic, resulting in embeddings that fail to capture fine semantic nuances. 
  
  \item \textbf{Interpretability Issue:} 
  A significant proportion of questions consistently yield simple \yes{} answers, which can make the resulting embeddings more challenging to interpret and analyze.
\end{enumerate}

These limitations reduce the effectiveness of the embeddings in capturing fine-grained semantic differences, which in turn hinders their practical utility in downstream tasks. 
To overcome these challenges, we introduce Contrastive Question Generation (CQG), a novel method that leverages the creative potential of LLMs to generate more discriminative questions.

\paragraph{Contrastive Question Generation (CQG)}
The CQG method applies contrastive learning principles, using \emph{positive}, \emph{hard negative}, and \emph{easy negative} samples to guide LLMs in generating high-quality questions.
These questions are designed to effectively differentiate positive samples from negative ones, especially hard negatives, which are semantically similar \citep{robinson2021contrastive}.
The goal is to generate questions that elicit a \yes{} answer only for a specific group of texts while excluding others, even those that are closely related.

\begin{example} \label{example:cqg}
  Consider a toy example with four groups of texts for animals: $\mathbb{G}_1=\{\TwoQ{Whale}, \TwoQ{Dolphin}\}$, $\mathbb{G}_2=\{\TwoQ{Shark}, \TwoQ{Ray}\}$, $\mathbb{G}_3=\{\TwoQ{Giraffe}, \TwoQ{Deer}\}$, and $\mathbb{G}_4=\{\TwoQ{Eagle}, \TwoQ{Hawk}\}$. 
  The objective is to generate questions that can effectively distinguish $\mathbb{G}_1$ from other groups. 
  At first, broad questions such as \Q{Does it live in water?} or \Q{Is it a mammal?} might seem useful. 
  While these questions correctly yield \yes{} for \TwoQ{Whale} and \TwoQ{Dolphin}, they also apply to other groups. 
  For example, \TwoQ{Shark} and \TwoQ{Ray} also live in water, and \TwoQ{Giraffe} and \TwoQ{Deer} are also mammals.
  To better differentiate $\mathbb{G}_1$, a more precise question would be \Q{Does the animal use echolocation to navigate and hunt?}, which yields a \yes{} only for \TwoQ{Whale} and \TwoQ{Dolphin}, effectively distinguishing them from the other groups.
  Furthermore, this question could also generalize to other animals, such as \TwoQ{Bat}, that were not part of the original groups, highlighting the method's potential to apply to unseen examples.
  \hfill $\triangle$ \par 
\end{example}

\begin{figure}[t]
  \centering
  \includegraphics[width=0.92\textwidth]{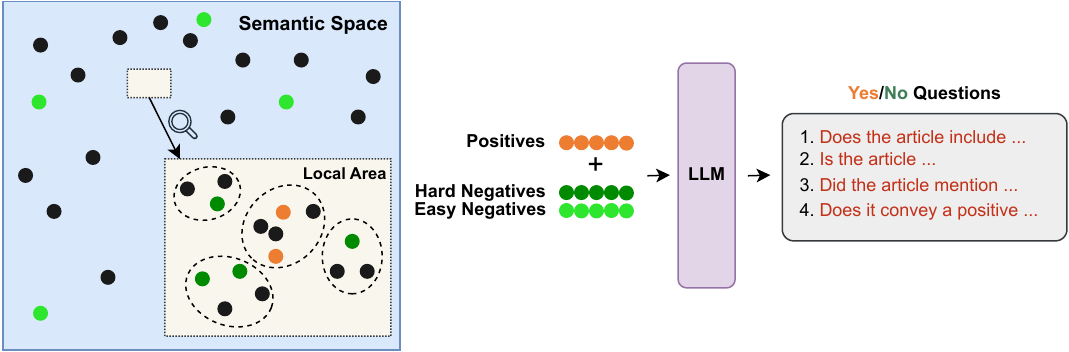}
  \vspace{-0.5em}
  \caption{Illustration of the CQG method.}
  \label{fig:contrastive-question-generation}
  \vspace{-0.5em}
\end{figure}

As depicted in Figure \ref{fig:contrastive-question-generation}, the CQG method begins by identifying semantically similar groups of texts. 
This is accomplished by encoding the text corpus into embedding vectors and clustering these vectors to form distinct groups.
For each cluster, we design a strategic sampling technique: selecting $n_{p}$ positive texts from within the cluster, $n_{h}$ hard negative samples from neighboring clusters, and $n_{e}$ easy negative samples from the global corpus. 
The LLM is then prompted to generate questions under a key constraint: the questions must elicit \yes{} answers exclusively for the positive samples and \no{} answers for all negative samples. The detailed prompt for CQG is provided in Appendix \ref{app:prompts:contrastive-question-generation}.

This strategic sampling technique serves two main purposes:
\vspace{-0.25em}
\begin{enumerate}[nolistsep,leftmargin=30pt,label*=(\arabic*)] 
  \item \textbf{Discriminative Power:} 
  Contrasting positive samples with hard negative samples encourages the LLM to generate highly discriminative questions tailored to each cluster. 
  
  \item \textbf{Broader Relevance:}
  Including easy negative samples ensures that the generated questions maintain broader relevance across the entire corpus. 
\end{enumerate}
A well-tuned negative-to-positive ratio, i.e., $(n_{h} + n_{e})/n_{p}$, encourages the LLM to craft precise, discriminative questions for the positive cluster, leading to sparser and more interpretable embedding dimensions.
This process is repeated across all clusters, resulting in a comprehensive set of LLM-generated questions that capture the unique characteristics of each cluster while maintaining global relevance, forming the foundation of our interpretable text embedding framework.

\paragraph{Post-Processing}
LLMs may encounter two common issues when generating questions: (1) failing to consistently provide \yes{} answers only for positive samples and (2) generating similar questions across different clusters.
To address these challenges, we implement a post-processing step to filter and select the highest-quality, non-redundant questions.

We introduce a probing mechanism to evaluate and refine the generated questions. 
For each question, we randomly sample $p_{p}$ positive probes from the originating cluster, $p_{h}$ hard negative probes from neighboring clusters, and $p_{e}$ easy negative probes from other clusters. 
The LLM answers these questions, and we calculate the quality of each question using the following formula:
\begin{equation}\label{eqn:quality}
  \text{quality} = \frac{\text{\# \yes{} for positive probes}}{p_{p}} - \frac{\text{\# \yes{} for negative probes}}{p_{h} + p_{e}}
\end{equation}
Equation \ref{eqn:quality} measures the difference between the percentage of \yes{} answers for positive probes and negative probes. 
A higher quality value indicates that the question is more discriminative for the cluster, as it correctly identifies more positive samples while filtering out negatives.

To construct the final set of questions, we iteratively select the top $t$ most discriminative questions from each cluster, based on their quality values, and ensure that no two questions are highly semantically similar.
Here, the similarity between questions is measured using cosine similarity between their corresponding embedding vectors, which are generated using a pre-trained encoding model. 
If the cosine similarity of two questions exceeds a predefined threshold $\theta$, they are considered duplicates, and the latter question is excluded.

This post-processing step helps filter out hallucinated and/or low-quality questions generated by the LLM. 
The final set comprises $m$ questions, forming a diverse and highly discriminative collection that effectively captures the semantic structure of the entire corpus.

\subsection{Question Answering}
\label{sect:framework:answer}

\paragraph{Motivations}
Generating answers to questions using LLMs can be prohibitively expensive, especially when scaling up to large datasets with numerous questions. 
Table \ref{tab:cost-analysis} presents the cost analysis for embedding the \data{MS MARCO} dataset \citep{bajaj2016ms}, revealing that the expense of LLM-generated answers becomes untenable when dealing with thousands of questions (dimensions) across million-scale datasets. 
LLM-based Question Answering (QA), which leverages LLMs to answer 10,000 questions for approximately 8.8 million articles in the \data{MS MARCO} dev set, requires about 4.4 billion LLM inference passes and processes 1.5 trillion tokens. This incurs a substantial cost of \textbf{244,551 USD}, even with a cost-effective model (\chkp{GPT-4o-mini}) and a token-efficient prompting approach (grouping 20 questions per prompt). 
Further details are available in Appendix \ref{app:cost-analysis}.

While increasing the number of questions improves performance (see Section \ref{sect:expt:number-of-questions}), the cost associated with LLM-based QA renders it impractical for large-scale real-world applications. 
To address this, we propose the Multi-task Binary Question Answering (MBQA) model as a cost-effective alternative.

\paragraph{Multi-task Binary Question Answering (MBQA) Model}
The MBQA model is designed to leverage LLM-generated answers from a smaller subset of texts to train a multi-task binary classification model.
This model consists of a pre-trained encoding model and multiple classification heads. The encoder converts the input text into an embedding vector, while the classification heads predict binary scores for each question.
Formally, the MBQA model $\mathcal{M}$ is defined as:
\begin{equation}
  \mathcal{M} = (Enc, \{C_1, C_2, \cdots, C_m\}),
\end{equation}
where $Enc: \mathcal{T} \rightarrow \mathbb{R}^d$ represents the encoding model, and $C_i: \mathbb{R}^d \rightarrow [0, 1]$ is the $i$-th Multi-Layer Perceptron (MLP) classification head.
For a given input text $\bm{t} \in \mathcal{T}$, the MBQA model generates a binary embedding vector $\bm{y} = [y_1, y_2, \cdots, y_m] \in \{0, 1\}^m$ as follows:
\begin{align}
  \bm{e} &= Enc(\bm{t}), \\
  y_i &= \bm{1}[\sigma(C_i(\bm{e})) > \tau],~\text{for}~i \in \{1, 2, \cdots, m\},
\end{align}
where $\sigma$ is the Sigmoid function, $\bm{1}[\cdot]$ is the indicator function, and $\tau$ is the threshold for binary classification.
During training, the encoder $Enc$ is frozen and only the classification heads $\{C_1, C_2, \cdots, C_m\}$ are optimized using weighted Binary Cross-Entropy (BCE) Loss \citep{bishop2006pattern} on the available LLM-generated question-answer pairs.

\paragraph{Remarks}
The MBQA model achieves 96\% accuracy in reproducing LLM-generated answers for CQG questions with just a single pass through the encoding model, substantially reducing costs compared to running a pre-trained LLM for each text. 
Our model only requires training data from as few as 1,000 articles per question, resulting in 10 million text-question pairs for 10,000 questions, costing just 31 USD using \chkp{GPT-4o-mini}. 
The training process takes 36 hours, and embedding the entire \data{MS MARCO} dev set requires 90 hours on a single GTX 1080 Ti, which is an inexpensive GPU. 
Consequently, encoding the same \data{MS MARCO} dev set with the MBQA model costs around \textbf{41 USD}--just \textbf{0.017\%} of the original cost with \chkp{GPT-4o-mini}. 
This model allows us to scale up the number of questions (dimensions) efficiently, providing interpretable embeddings at a fraction of the cost of LLM-based answering. For more details on training and evaluation, see Appendix \ref{app:question-answering-performance}.

\begin{table}[t]
\centering
\footnotesize
\caption{Estimated cost for embedding the \data{MS MARCO} dev set using LLM-generated answers.}
\label{tab:cost-analysis}
\vspace{-0.5em}
\begin{tabular}{lrrrrr}
  \toprule
  \multirow{2}{*}{\textbf{Model}} & \multicolumn{5}{c}{\textbf{Cost for Number of Questions}} \\ \cmidrule(lr){2-6}
  & 2,000 & 4,000 & 6,000 & 8,000 & 10,000 \\
  \midrule
  \textbf{GPT-4o-mini} & \$48,859    & \$97,792   & \$146,725   & \$195,570    & \$244,551    \\
  \textbf{GPT-4o}      & \$1,454,000 & \$2,910,487 & \$4,366,946 & \$5,820,517 & \$7,278,566 \\
  \cmidrule{1-6}
  \textbf{MBQA}        & \$13 & \$20 & \$27 & \$34 & \$41 \\
  \bottomrule
\end{tabular}
\end{table}
\setlength{\textfloatsep}{1.5em}

\section{Experiments}
\label{sect:expt}

In this section, we present a comprehensive evaluation of the \algo{CQG-MBQA} framework by addressing four essential questions aimed at understanding its performance and applicability:
\begin{itemize}[nolistsep,leftmargin=25pt]
  \item \textbf{Embedding Quality:} How well does our framework generate high-quality interpretable embeddings comparable to advanced black-box models? (Section \ref{sect:expt:main:embedding-quality})
  
  \item \textbf{Interpretability:} Does our framework improve the human interpretability of embeddings over existing methods? (Section \ref{sect:expt:main:interpretability})
  
  \item \textbf{Question Efficiency:} Can the CQG method generate a limited number of highly discriminative questions that maintain strong performance? (Section \ref{sect:expt:number-of-questions})
  
  \item \textbf{Flexibility:} Can the MBQA model be tuned to strike a balance between embedding quality and interpretability? (Section \ref{sect:expt:embedding-quality-interpretability-trade-off})
\end{itemize}
To rigorously evaluate the framework, we conduct experiments on three core downstream tasks in text embedding: STS, retrieval, and clustering. These experiments allow us to benchmark the \algo{CQG-MBQA} framework against both black-box and interpretable models.

\subsection{Metrics}
\label{sect:expt:metrics}

\paragraph{Embedding Quality Measurement}
For evaluating embedding quality, we adopt the metrics that are widely used in the MTEB benchmark \citep{muennighoff-etal-2023-mteb} for a comprehensive comparison. 
For STS tasks, we use Spearman correlation \citep{spearman1904proof} on cosine similarity between embeddings as the evaluation metric. In retrieval tasks, we assess the performance using Normalized Discounted Cumulative Gain at Top 10 (nDCG@10) \citep{wang2013theoretical}. For clustering tasks, we evaluate the results using V-Measure \citep{rosenberg-hirschberg-2007-v}.

\paragraph{Interpretability Measurement}
Since both STS and retrieval tasks measure pairwise text similarity using cosine similarity, we focus on interpreting the cosine similarity scores produced by \algo{CQG-MBQA}. 
With inherently interpretable dimensions, we can offer insights to users by highlighting the dimensions that contribute most to the similarity between two texts.
Building on \algo{COGAM} \citep{abdul2020cogam}, we suggest that interpretability should account for the \textbf{cognitive load} imposed on users. 
In \algo{COGAM}, cognitive load is assessed by counting the number of visual cognitive chunks.
Similarly, we measure it by the number of questions a user must consider to understand the similarity between two texts, corresponding to the dimensions where both embedding vectors have a value of 1. 
Formally, for any two binary embedding vectors $\bm{u}=[u_1,u_2,\cdots,u_m]$ and $\bm{v}=[v_1,v_2,\cdots,v_m]$, the cognitive load is defined as the inner product of $\bm{u}$ and $\bm{v}$:
\begin{equation}\label{eqn:cognitive-load}
  \text{cognitive load} = \ip{\bm{u}, \bm{v}} = \sum_{i=1}^m u_i v_i.
\end{equation}
Quantifying cognitive load allows us to assess the interpretability of our \algo{CQG-MBQA} framework's embeddings. 
A lower value indicates that fewer dimensions are involved, making the interpretation easier to understand, thus enhancing both interpretability and user-friendliness.

\begin{table}[t]
\centering
\small
\caption{STS results measured by Spearman correlation. Evaluated on seven popular datasets: \data{SemEval STS tasks 2012-2016 (STS12--STS16)} \citep{agirre-etal-2012-semeval, agirre-etal-2013-sem, sts14, sts15, sts16}, \data{STS Benchmark (STS-B)} \citep{sts-benchmark}, and \data{SICK-Relatedness (SICK-R)} \citep{sick-r} using the MTEB evaluation suite \citep{muennighoff-etal-2023-mteb}.}
\label{tab:sts-results}
\vspace{-0.5em}
\resizebox{0.99\textwidth}{!}{
\begin{tabular}{cccccccccc}
  \toprule
  \multirow{2}{*}{\textbf{Type}}&\multirow{2}{*}{\textbf{Model}} & \multicolumn{8}{c}{\textbf{Spearman Correlation (STS)}}  \\
  \cmidrule(lr){3-10} 
  & & \data{STS12}  & \data{STS13} &  \data{STS14} & \data{STS15}  & \data{STS16}  & \data{STS-B} & \data{SICK-R} & \data{Avg.} \\
  \midrule
  \multirow{10}{*}{\rotatebox{90}{\textbf{Black-box}}} 
  & \algo{BERT}            & 38.78 & 57.98 & 57.98 & 63.15 & 61.06 & 46.35 & 58.40 & 54.81 \\
  & \algo{GloVe}           & 54.64 & 69.16 & 60.81 & 72.31 & 65.34 & 61.54 & 55.43 & 62.74 \\
  & \algo{USE}             & 64.49 & 67.80 & 64.61 & 76.83 & 73.18 & 74.92 & 76.69 & 71.22 \\
  & \algo{SimCSE (Unsup.)} & 66.05 & 81.49 & 73.61 & 79.72 & 78.12 & 76.52 & 72.24 & 75.39 \\
  & \algo{SBERT (Ori.)}    & 74.53 & 77.00 & 73.18 & 81.85 & 76.82 & 79.10 & 74.29 & 76.68 \\
  & \algo{SimCSE (Sup.)}   & 75.30 & 84.67 & 80.19 & 85.40 & 80.82 & 84.25 & 68.38 & 79.86 \\
  & \algo{WhitenedCSE}     & 74.65 & 85.79 & 77.49 & 84.71 & 80.33 & 81.48 & 75.34 & 79.97 \\
  & \algo{SBERT (New)}     & 73.08 & 82.13 & 76.73 & 85.58 & 80.23 & 83.09 & 79.32 & 80.02 \\
  & \algo{OpenAI}          & 72.84 & 86.1  & 81.15 & 88.49 & 85.08 & 83.56 & 79.00 & 82.31 \\
  & \algo{AnglE}           & 79.09 & 89.62 & 85.02 & 89.51 & 86.61 & 89.06 & 82.62 & 85.93 \\
  \cmidrule{1-10}
  \multirow{3}{*}{\rotatebox{90}{\textbf{Interp.}}} 
  & \algo{Bag-of-Tokens}   & 44.75 & 52.06 & 54.78 & 68.65 & 60.59 & 54.85 & 57.87 & 56.22 \\
  & \algo{QAEmb-MBQA}      & 59.40 & 63.19 & 57.68 & 69.29 & 63.18 & 71.33 & 72.33 & 65.20 \\
  & \algo{CQG-MBQA}        & 69.21 & 80.19 & 73.91 & 80.66 & 78.30 & 82.69 & 78.21 & 77.60 \\
  \bottomrule
\end{tabular}}
\end{table}

\begin{table}[t]
\centering
\small
\caption{Retrieval results evaluated by nDCG@10. Evaluated on seven diverse datasets: \data{MS MARCO} \citep{bajaj2016ms}, \data{NewsSpectrum (NSP)} \citep{sun2024diversinews}, \data{ArguAna} \citep{wachsmuth-etal-2018-retrieval}, \data{FiQA-2018 (FQA)} \citep{fiqa2018}, \data{NFCorpus (NFC)} \citep{boteva2016full}, \data{SCIDOCS} \citep{cohan-etal-2020-specter}, and \data{SciFact} \citep{wadden-etal-2020-fact}. 
\data{MS MARCO} is evaluated on a 1\% sample of its dev set, while \data{NewsSpectrum} uses news titles as queries with corresponding articles as targets. The remaining datasets are assessed using the MTEB evaluation suite.}
\label{tab:retrieval-results}
\vspace{-0.5em}
\resizebox{0.99\textwidth}{!}{
\begin{tabular}{cccccccccc}
  \toprule
  \multirow{2}{*}{\textbf{Type}}&\multirow{2}{*}{\textbf{Model}} & \multicolumn{8}{c}{\textbf{nDCG@10 (Retrieval)}}  \\
  \cmidrule(lr){3-10} 
  && \data{MS MARCO}  & \data{NSP} & \data{ArguAna} & \data{FQA}  & \data{NFC}  & \data{SCIDOCS} & \data{SciFact} & \data{Avg.} \\
  \midrule
  \multirow{7}{*}{\rotatebox{90}{\textbf{Black-box}}}
  & \algo{BERT}            & 16.86 & 10.64 & 28.29 & 2.19  & 4.30  & 2.82  & 13.34 & 11.21 \\
  & \algo{SimCSE (Unsup.)} & 44.63 & 38.04 & 38.34 & 9.84  & 9.88  & 5.50  & 25.72 & 24.56 \\
  & \algo{GloVe}           & 44.27 & 30.84 & 36.30 & 10.09 & 13.87 & 8.04  & 29.58 & 24.71 \\
  & \algo{SimCSE (Sup.)}   & 47.86 & 46.29 & 38.33 & 10.41 & 12.42 & 7.53  & 29.59 & 27.49 \\
  & \algo{SBERT (New)}     & 88.74 & 70.64 & 47.13 & 37.27 & 32.25 & 21.82 & 62.64 & 51.50 \\
  & \algo{AnglE}           & 90.43 & 82.87 & 66.15 & 44.84 & 38.65 & 22.98 & 74.07 & 60.00 \\
  & \algo{OpenAI}          & 92.18 & 86.77 & 58.05 & 55.00 & 42.07 & 23.11 & 77.77 & 62.14 \\
  \cmidrule{1-10}
  \multirow{4}{*}{\rotatebox{90}{\textbf{Interp.}}} 
  & \algo{Bag-of-Tokens}   & 29.79 & 23.95 & 34.25 & 3.99  & 21.51 & 6.79  & 47.36 & 23.95 \\
  & \algo{BM25}            & 68.42 & 76.42 & 49.28 & 25.14 & 32.08 & 15.78 & 68.70 & 47.97 \\
  & \algo{QAEmb-MBQA}      & 40.51 & 16.07 & 34.75 & 8.23  & 3.87  & 3.74  & 12.01 & 17.03 \\
  & \algo{CQG-MBQA}        & 62.21 & 32.30 & 47.75 & 18.63 & 9.74  & 8.67  & 32.80 & 30.30 \\
  \bottomrule
\end{tabular}}
\end{table}

\begin{table}[t]
\centering
\caption{Clustering results assessed by V-Measure. Evaluated on seven commonly-used datasets: \data{TwentyNewsgroupsClustering (TNG)}, \data{StackExchangeClusteringP2P (SE-P2P)}, \data{BiorxivClusteringP2P (BR-P2P)}, \data{BiorxivClusteringS2S (BR-S2S)}, \data{MedrxivClusteringP2P (MR-P2P)}, \data{MedrxivClusteringS2S (MR-S2S)}, and \data{RedditClusteringP2P (RD-P2P)} from the MTEB evaluation suite.}
\label{tab:clustering-results}
\vspace{-0.5em}
\resizebox{0.99\textwidth}{!}{
\begin{tabular}{cccccccccc}
  \toprule
  \multirow{2}{*}{\textbf{Type}}&\multirow{2}{*}{\textbf{Model}} & \multicolumn{7}{c}{\textbf{V-Measure (Clustering)}}  \\
  \cmidrule(lr){3-10} 
  && \data{TNG}  & \data{SE-P2P} & \data{BR-P2} & \data{BR-S2S}  & \data{MR-P2P}  & \data{MR-S2S} & \data{RD-P2P} & \data{Avg.} \\
  \midrule
  \multirow{7}{*}{\rotatebox{90}{\textbf{Black-box}}}
  & \algo{SimCSE (Unsup.)} & 23.21 & 28.50 & 24.90 & 19.55 & 23.60 & 21.97 & 45.14 & 26.70 \\
  & \algo{GloVe}           & 25.83 & 28.51 & 29.27 & 19.18 & 26.12 & 20.38 & 35.82 & 26.44 \\
  & \algo{BERT}            & 23.35 & 26.55 & 30.12 & 24.77 & 26.09 & 23.60 & 43.32 & 28.26 \\
  & \algo{SimCSE (Sup.)}   & 34.86 & 29.45 & 30.15 & 24.67 & 26.25 & 24.12 & 47.74 & 31.03 \\
  & \algo{SBERT (New)}     & 47.47 & 33.13 & 36.99 & 33.21 & 34.25 & 32.24 & 54.80 & 38.87 \\
  & \algo{AnglE}           & 51.72 & 36.72 & 39.38 & 37.23 & 33.22 & 31.18 & 65.35 & 42.11 \\
  & \algo{OpenAI}          & 58.14 & 36.88 & 38.03 & 36.53 & 32.70 & 31.27 & 67.96 & 43.07 \\
  \cmidrule{1-10}
  \multirow{3}{*}{\rotatebox{90}{\textbf{Interp.}}} 
  & \algo{Bag-of-Tokens}   & 8.52 & 17.64 & 4.70 & 3.32 & 11.39 & 13.05 & 15.67 & 10.61 \\
  & \algo{QAEmb-MBQA}      & 36.72 & 25.68 & 24.66 & 21.16 & 25.53 & 22.85 & 46.57 & 29.02 \\
  & \algo{CQG-MBQA}        & 40.00 & 28.22 & 34.88 & 31.13 & 31.02 & 28.71 & 54.40 & 35.48 \\
  \bottomrule
\end{tabular}}
\end{table}

\subsection{Models}
\label{sect:expt:models}

\paragraph{Interpretable Models}
We evaluate \algo{CQG-MBQA} against existing interpretable baselines to provide a thorough comparison. 
The implementation details of \algo{CQG-MBQA} are provided in Appendix \ref{app:experiment-details:cqg-mbqa}.
To make a fair comparison and highlight the benefits of our CQG method, we introduce \algo{QAEmb-MBQA}. We replicate the example-based question generation approach of QAEmb \citep{Benara2024CraftingIE} and integrate the QAEmb-generated questions with our MBQA model. 
Additionally, we include \algo{Bag-of-Tokens}, a simple baseline that uses the BERT tokenizer to produce interpretable embeddings.
For the retrieval task, we also compare against the rule-based sparse retriever \algo{BM25} \citep{robertson1995okapi}, implemented using BM25S \citep{lù2024bm25sordersmagnitudefaster}.

\paragraph{Black-box Models}
To benchmark the embedding quality of \algo{CQG-MBQA}, we compare it with several advanced black-box text embedding models. These include 
\algo{GloVe} \citep{pennington-etal-2014-glove, reimers-gurevych-2019-sentence}, \algo{USE} \citep{cer-etal-2018-universal}, \algo{BERT} \citep{devlin-etal-2019-bert}, the \algo{Original (Ori.)} and \algo{Up-to-date (New)} versions of \algo{Sentence-BERT (SBERT)} \citep{reimers-gurevych-2019-sentence}, the \algo{Supervised (Sup.)} and \algo{Unsupervised (Unsup.)} versions of \algo{SimCSE} \citep{gao-etal-2021-simcse}.
Additionally, we compare our framework with \algo{WhitenedCSE} \citep{zhuo-etal-2023-whitenedcse}, the \algo{OpenAI} API, and \algo{AnglE} \citep{li-li-2024-aoe}.
Implementation details for all baseline models are outlined in Appendix \ref{app:experiment-details:baselines}.

\subsection{Embedding Quality}
\label{sect:expt:main:embedding-quality}

Tables \ref{tab:sts-results}--\ref{tab:clustering-results} present the embedding quality results for the STS, retrieval, and clustering tasks, highlighting \algo{CQG-MBQA}'s competitive performance.
In the STS tasks (Table \ref{tab:sts-results}), \algo{CQG-MBQA} achieves comparable embedding quality to advanced dense embedding models like \algo{SimCSE} and \algo{SBERT (New)} while preserving inherent interpretability. 
It outperforms earlier methods such as \algo{GloVe}, \algo{USE}, and \algo{BERT}, as well as all interpretable baselines.
For retrieval tasks (Table \ref{tab:retrieval-results}), \algo{CQG-MBQA} surpasses models like \algo{SimCSE}, \algo{GloVe}, and \algo{BERT}. 
While it trails state-of-the-art black-box models such as \algo{AnglE} and \algo{OpenAI}, it consistently outperforms all interpretable baselines except for \algo{BM25}, which is a rule-based model optimized for retrieval.
In clustering tasks (Table \ref{tab:clustering-results}), \algo{CQG-MBQA} outperforms several black-box models (\algo{SimCSE}, \algo{GloVe}, and \algo{BERT}) and surpasses all interpretable baselines. 
Its performance closely approaches that of more recent models like \algo{SBERT (New)}.
Additionally, the comparison between \algo{CQG-MBQA} and \algo{QAEmb-MBQA} underscores the efficacy of our CQG algorithm in generating high-quality, discriminative questions that capture semantic nuances.

\subsection{Interpretability}
\label{sect:expt:main:interpretability}

\paragraph{Cognitive Load}
Table \ref{tab:cognitive-load} displays the cognitive load required to interpret embeddings across different interpretable models, evaluated through the STS task, which computes pairwise similarity of texts. 
This provides a clear measure of how much effort is needed to understand the embeddings. 
\algo{CQG-MBQA} achieves a 2.5$\sim$3.6$\times$ lower cognitive load than \algo{QAEmb-MBQA}, indicating that the CQG method significantly enhances interpretability. 
This technique produces more \no{} answers and fewer \yes{} answers, making the embeddings easier to interpret.
For \algo{Bag-of-Tokens}, it has a much lower cognitive load due to its lexical nature, but this advantage comes at the cost of significantly reduced embedding quality. 
The trade-off between interpretability and embedding quality can be adjusted by tuning the binary classification threshold $\tau$, as further discussed in Section \ref{sect:expt:embedding-quality-interpretability-trade-off}.

\begin{table}[t]
\centering
\footnotesize
\caption{STS results measured by cognitive load (the lower the better).}
\label{tab:cognitive-load}
\vspace{-0.5em}
\resizebox{0.99\textwidth}{!}{
\begin{tabular}{ccccccccc}
  \toprule
  \multirow{2}{*}{\textbf{Model}} & \multicolumn{8}{c}{\textbf{Cognitive Load (STS)}}  \\
  \cmidrule(lr){2-9} 
  & \data{STS12}  & \data{STS13} &  \data{STS14} & \data{STS15}  & \data{STS16}  & \data{STS-B} & \data{SICK-R} & \data{Avg.} \\
  \midrule
  \algo{Bag-of-Tokens} & 8     & 4     & 6     & 5     & 8     & 7     & 6     & 6     \\
  \algo{QAEmb-MBQA}    & 1,626 & 1,571 & 1,625 & 1,443 & 1,577 & 1,408 & 1,018 & 1,467 \\
  \algo{CQG-MBQA}      & 481   & 439   & 458   & 426   & 478   & 446   & 413   & 449   \\
  \bottomrule
\end{tabular}}
\end{table}

\begin{figure}[t]
  \centering
  \includegraphics[width=0.99\textwidth]{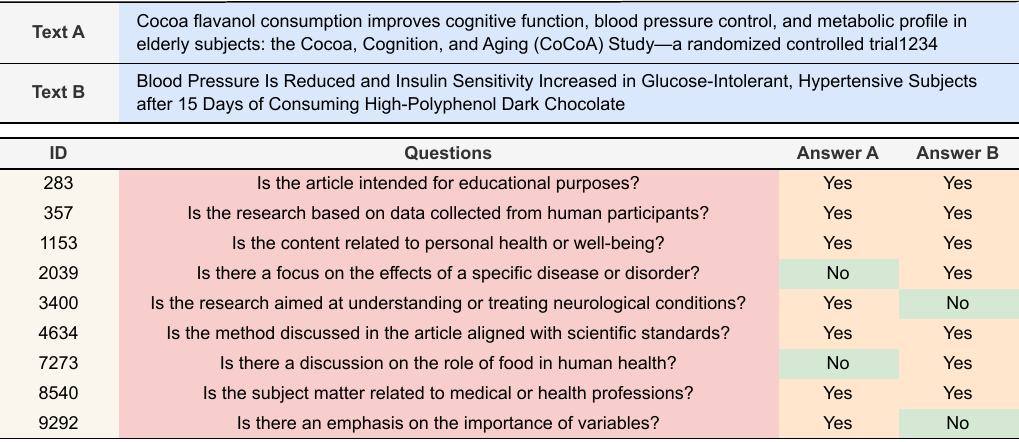}
  \vspace{-0.5em}
  \caption{Case study.}
  \label{fig:case-study}
  \vspace{-0.5em}
\end{figure}

\begin{figure}[t]
  \centering
  \includegraphics[width=0.99\textwidth]{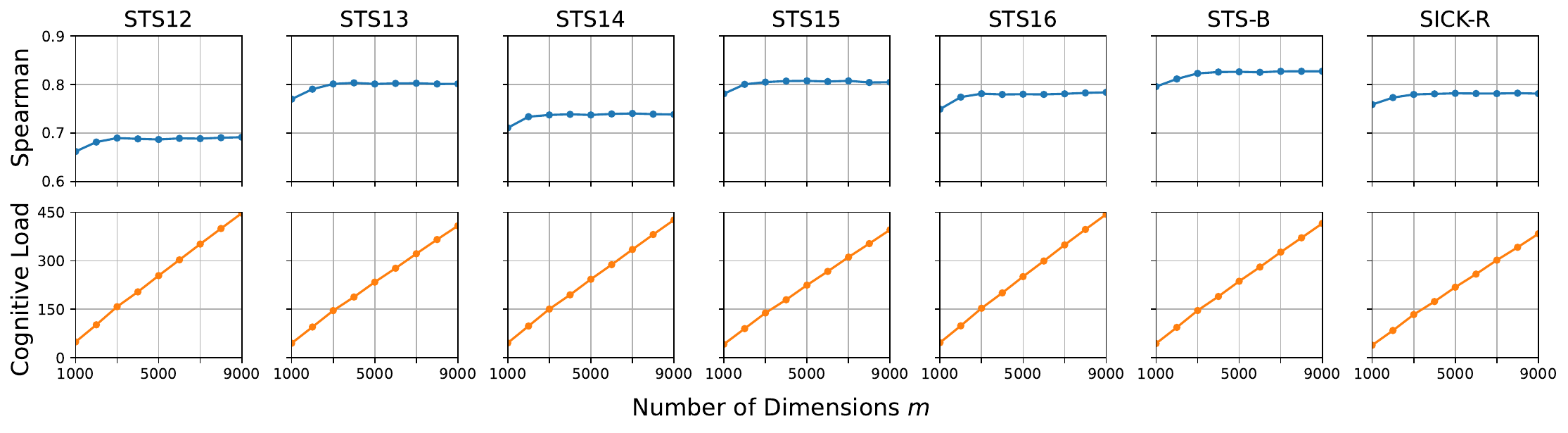}
  \vspace{-0.6em}
  \caption{Spearman correlation and cognitive load vs.~the number of dimensions $m$. Higher Spearman correlation signals better embedding quality; lower cognitive load implies greater interpretability.}
  \label{fig:abl-dimensions}
\end{figure}

\begin{figure}[t]
  \centering
  \includegraphics[width=0.99\textwidth]{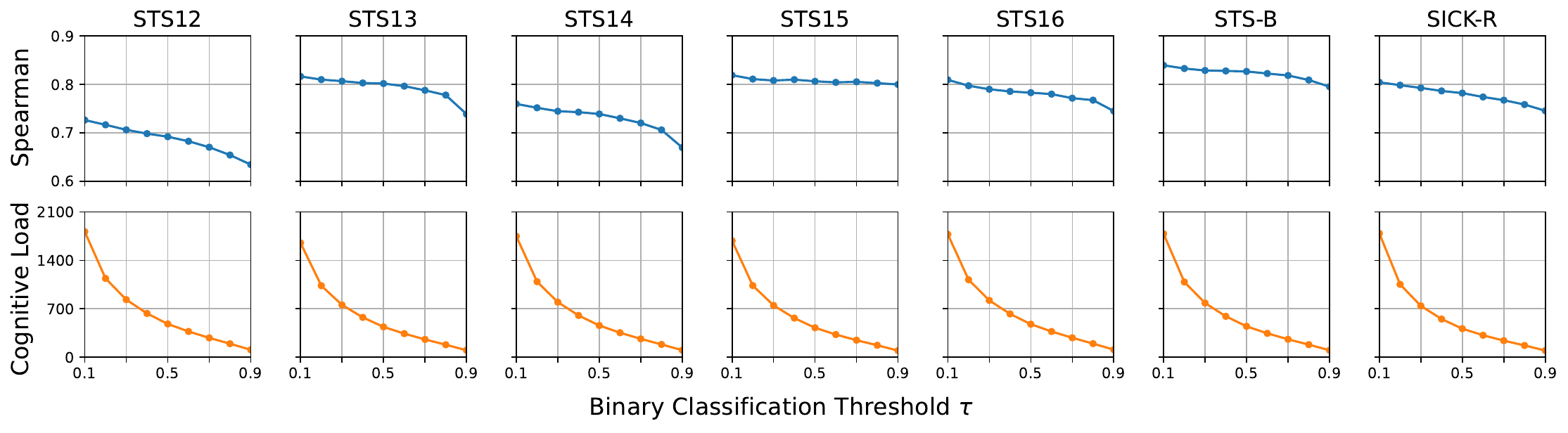}
  \vspace{-0.6em}
  \caption{Spearman correlation and cognitive load vs.~the binary classification threshold $\tau$.}
  \label{fig:abl-thresholds}
\end{figure}

\paragraph{Case Study}
Figure \ref{fig:case-study} showcases a pair of texts from our training corpus, focusing on nine specific questions (dimensions) where at least one text yield a \yes{} answer. 
This illustrates how \algo{CQG-MBQA} generates relevant and insightful dimensions.
For instance, question ID 1153, which asks if the text is related to personal health or well-being, receives a \yes{} for both Text A and Text B, accurately reflecting their shared focus on health topics.
Similarly, question ID 4634 inquires whether the text aligns with scientific standards, and both texts--discussing evidence-based findings on substance effects--obtain \yes{} answers, showcasing the relevance of generated questions.

Despite the texts' similarity, \algo{CQG-MBQA} captures subtle semantic differences through fine-grained questions. 
For example, question ID 3400 asks if the research targets neurological conditions. 
Text A, which discusses cognitive function, receives a \yes{}, indicating a connection to neurological conditions, whereas Text B, focusing on blood pressure and insulin sensitivity, acquires a \no{}, highlighting a clear distinction in their subject matter.
This case study highlights how CQG generates interpretable, relevant, and discriminative questions that effectively capture nuanced semantic differences, while MBQA accurately predicts the answers, reinforcing the framework's practicality and reliability.

\subsection{Effect on Number of Questions}
\label{sect:expt:number-of-questions}

We explore how varying the number of questions (dimensions) $m$ impacts the quality and interpretability of embeddings produced by \algo{CQG-MBQA}. 
To adjust $m$, we reduce the output length of the final binary embedding vector.
Figure \ref{fig:abl-dimensions} illustrates the relationship among embedding quality, interpretability, and $m$. 
As $m$ grows, the Spearman correlation increases and stabilizes around 3,000 dimensions, indicating better embedding quality. 
This highlights the need for an efficient QA model to manage computational costs while scaling up dimensions for optimal embedding quality.

However, a trade-off arises: as $m$ increases, cognitive load increases, and the interpretability declines due to a higher proportion of \yes{} in the embeddings. 
This inverse relationship between embedding quality and interpretability emphasizes the importance of balancing dimensions based on the task’s requirements.
Figure \ref{fig:abl-dimensions} also demonstrates the effectiveness of the CQG algorithm in generating high-quality, discriminative embeddings with approximately 3,000 questions, achieving a balance between embedding quality and interpretability without an excessive number of questions.

\subsection{Trade-off between Embedding Quality and Interpretability}
\label{sect:expt:embedding-quality-interpretability-trade-off}

To further investigate the balance between embedding quality and interpretability, we vary the binary classification threshold $\tau$ that determines the final binary embedding vector.
Figure \ref{fig:abl-thresholds} depicts a clear trade-off between embedding quality and interpretability. 
Increasing $\tau$ improves the interpretability, but this comes at the cost of reduced embedding quality, as the Spearman correlation decreases. 
This is due to fewer active dimensions, leading to reduced values in the cognitive load, which are easier to interpret but may lose subtle semantic distinctions.

More importantly, this trade-off presents an opportunity for user-driven customization. 
Depending on different scenarios, users of our framework can dynamically tune the desired $\tau$ based on the cognitive load to meet their needs. 
For instance, in scenarios requiring rapid decision-making or where cognitive resources are constrained, users might prioritize interpretability by opting for a higher threshold. 
On the other hand, in situations where nuanced analysis is crucial and resources are abundant, a lower threshold could be chosen to maximize embedding quality. 
This flexibility makes \algo{CQG-MBQA} highly adaptable to different scenarios and user requirements.

\section{Conclusion}
\label{sec:conclusion}

In this paper, we introduce \algo{CQG-MBQA}, a novel general framework for generating interpretable semantic text embeddings by systematically creating binary questions and using the answers as interpretable embedding dimensions. 
Our CQG method effectively addresses the challenges of generalizability and quality issues during the question generation phase, while the MBQA model provides an efficient, scalable solution for answering these questions, significantly reducing costs. 
Through extensive experiments on STS, retrieval, and clustering tasks, we demonstrate that our framework delivers performance comparable to advanced black-box models while being inherently interpretable. 
Moreover, \algo{CQG-MBQA} consistently outperforms other interpretable text embedding models across various downstream tasks, further validating its effectiveness.




\bibliography{iclr2025_conference}
\bibliographystyle{iclr2025_conference}

\appendix

\section{Prompts}
\label{app:prompts}

\subsection{Prompt: Contrastive Question Generation}
\label{app:prompts:contrastive-question-generation}

We present the prompt template used for the Contrastive Question Generation (CQG) algorithm. The input to the prompt template consists of two lists of texts: \textit{positive\_examples} and \textit{negative\_examples}. The LLM is explicitly instructed to generate questions that yield \yes{} answers for the positive examples and \no{} answers for the negative examples.
To enhance the quality of the generated questions and the potential generalizability to new texts, we prompt the LLM to avoid complex sentence structures. Initial experiments revealed that if we do not require the LLM to avoid complex sentence structures, the LLM tended to create discriminative questions by simply combining two conditions to ensure \yes{} answers for the positives, rather than identifying deeper relationships between the positives and negatives. 
Additionally, we include formatting instructions at the end of the prompt template to improve result parsing accuracy.

\begin{tcolorbox}
  Generate 10 simple yet insightful yes/no questions that determine the properties of an article, where for all questions, the answer will be ``yes'' for ALL the positive articles and ``no'' for ALL the negative articles. Keep questions concise and avoid using complex sentence structures with ``and'' or ``or'' unless necessary.

  \textbf{Positive Articles:} \\
  Positive \{\textit{i}\}. \{\textit{positive\_example\_i}\}

  \textbf{Negative Articles:} \\
  Negative \{\textit{i}\}. \{\textit{negative\_example\_i}\}

  \textbf{Instruction:} Based on the excerpts provided, generate 10 simple yet insightful yes/no questions that can accurately differentiate the positive articles from the negative articles. Each question should be concise and framed in such a way that it will elicit a ``yes'' response for ALL positive articles and a ``no'' response for ALL negative articles. Avoid using complex sentence structures with ``and'' or ``or'' unless absolutely necessary. Format the questions in a numbered list as shown below:\\
  1. First simple yes/no question\\
  2. Second simple yes/no question
\end{tcolorbox}

\subsection{Prompt: QAEmb Question Generation}
\label{app:prompts:qaemb-question-generation}

The Question Generation Prompt \#5 used in the QAEmb paper \citep{Benara2024CraftingIE} is the most general form prompt in the list, making it suitable for adopting it to generate questions for general-purpose text embedding. It uses two lists of texts as input: \textit{example narrative sentences} and \textit{example yes/no questions}. The original prompt instructs the LLM to \textit{"Generate a bulleted list of 100 specific, non-overlapping yes/no questions that ask about aspects of the example narrative sentences that are important for classifying them."} This was originally designed for the task of fMRI prediction with narrative sentences.

Following this approach, we designed a prompt template for experiments on QAEmb question generation, also using two lists of texts as inputs: \textit{reference\_articles} and \textit{example\_questions}. The example questions are sourced from the original QAEmb paper, while the reference articles are randomly drawn from the training dataset. 

\begin{tcolorbox}
  Generate 10 diverse insightful yes/no questions that determine the properties of an article.

  \textbf{Reference Articles:} \\
  \{\textit{i}\}. \{\textit{reference\_article\_i}\}

  \textbf{Example Questions:} \\
  \{\textit{i}\}. \{\textit{example\_question\_i}\}

  \textbf{Instruction:} Based on the excerpts provided, generate 10 yes/no questions that can determine the properties of the articles. Format the questions in a numbered list as shown below:\\
  1. First yes/no question\\
  2. Second yes/no question
\end{tcolorbox}

\subsection{Prompt: Multi-task Binary Question Answering}
\label{app:prompts:question-answering}

This section details the prompt template used to generate LLM answers for training the Multi-task Binary Question Answering (MBQA) model. The prompt takes two inputs: the \textit{text\_chunk}, which is the training article sample, and a list of questions to be answered. To optimize token usage for the article sample and instructions, we group up to 20 questions in a single prompt.

\begin{tcolorbox}
  Evaluate the following text chunk based on the yes/no questions provided.

  \textbf{Text Chunk:}\\
  \{\textit{text\_chunk}\}

  \textbf{Questions:}\\
  \textit{i}. \{\textit{question\_i}\}

  \textbf{Instruction for the model:} Please read the provided text chunk and answer each of the questions with either "yes" or "no". Format the responses as follows: \\
  1. yes/no \\
  2. yes/no 
\end{tcolorbox}

\section{Training and Evaluation of the MBQA Model}
\label{app:question-answering-performance}

To ensure that the MBQA model produces faithful answers to the questions, we evaluate its question-answering performance on a 10\% held-out document set that was not used for training.

\paragraph{Data Collection}
For each question generated in the previous question generation step, we randomly sample 500 in-cluster samples, 300 neighboring cluster samples from 5 nearest clusters, and 200 random samples from the entire corpus. 
We use the pre-trained LLM (\chkp{GPT-4o-mini}) to generate answers for each question across these samples. The LLM-generated answers are batched in groups of 20 questions to train the multi-task binary classification model. 
Refer to Appendix \ref{app:prompts:question-answering} for the prompt used to collect answers. 
This approach allows us to gather data from a larger number of text samples, thereby increasing the generalizability of our model.

\paragraph{Training}
We train the MBQA model using the Adam optimizer with a learning rate $\alpha$ of 1e-4 and a batch size of one text sample. 
For each step, we calculate the loss based on all available questions with answers from the previous data collection phase. The model is trained using the BCEWithLogitsLoss function, where the weight is the ratio of \yes{} answers to \no{} answers in the training data.\footnote{\url{https://pytorch.org/docs/stable/generated/torch.nn.BCEWithLogitsLoss.html}} 
Specifically, for \algo{CQG-MBQA}, we set this weight to 7.5127, and for \algo{QAEmb-MBQA}, the weight is set to 4.9608. 
The model is trained for 3 million steps, at which point performance begins to converge.

\paragraph{Evaluation}
The classification results (with threshold $\tau=0.5$) on the held-out test set for \algo{CQG-MBQA} and \algo{QAEmb-MBQA} are presented in Tables \ref{tab:question-answering-performance-cqg} and \ref{tab:question-answering-performance-qaemb}, respectively. 
The \algo{CQG-MBQA} model achieves an accuracy of 96\% and a macro F1 score of 91\%, while the \algo{QAEmb-MBQA} model attains a high 93\% accuracy and 89\% macro F1 score. 
These results demonstrate that MBQA can accurately predict LLM-generated question answers, serving as a cost-effective substitute for the more expensive LLM model in generating embeddings.

\begin{table}[h]
\centering
\caption{Question answering performance of the \algo{CQG-MBQA} model.}
\label{tab:question-answering-performance-cqg}
\vspace{-0.5em}
\begin{tabular}{ccccc}
  \toprule
  \textbf{Class} & \textbf{Precision} & \textbf{Recall} & \textbf{F1-score} & \textbf{Support} \\
  \midrule
  \no          & 1.00 & 0.96 & 0.97 & 846,089 \\
  \yes         & 0.74 & 0.97 & 0.84 & 112,645 \\
  Macro Avg    & 0.87 & 0.96 & 0.91 & 958,734 \\
  Weighted Avg & 0.97 & 0.96 & 0.96 & 958,734 \\
  \cmidrule{1-5}
  Accuracy & \multicolumn{4}{c}{0.96} \\
  \bottomrule
\end{tabular}
\end{table}
\begin{table}[h]
\centering
\caption{Question answering performance of the \algo{QAEmb-MBQA} model.}
\label{tab:question-answering-performance-qaemb}
\vspace{-0.5em}
\begin{tabular}{ccccc}
  \toprule
  \textbf{Class} & \textbf{Precision} & \textbf{Recall} & \textbf{F1-score} & \textbf{Support} \\
  \midrule
  \no          & 0.99 & 0.93 & 0.96 & 886,749 \\
  \yes         & 0.72 & 0.94 & 0.82 & 178,591\\
  Macro Avg    & 0.85 & 0.93 & 0.89 & 1,065,340\\
  Weighted Avg & 0.94 & 0.93 & 0.93 & 1,065,340\\
  \cmidrule{1-5}
  Accuracy & \multicolumn{4}{c}{0.93} \\
  \bottomrule
\end{tabular}
\end{table}

\section{Cost Analysis for LLM-based QA vs. MBQA}
\label{app:cost-analysis}
We estimate the cost of LLM-based QA and MBQA for producing question answers for interpretable embeddings for the entire \data{MS MARCO} dev set. 

\paragraph{LLM-based QA}
Using LLMs to answer questions for document embedding is prohibitively expensive. Table \ref{tab:cost-analysis} shows the cost of running LLM-based QA on the \data{MS MARCO} dev set for various numbers of questions across different models. 
We assume grouping 20 questions into one prompt, using the format in Appendix \ref{app:prompts:question-answering}. 
Using the Batch API, the cost per 1 million tokens for \chkp{GPT-4o} is 2.5 USD for input tokens and 7.5 USD for output tokens, while for \chkp{GPT-4o-mini}, it's 0.075 USD for input tokens and 0.3 USD for output tokens.\footnote{Costs obtained from~\url{https://openai.com/api/pricing}}

\paragraph{MBQA}
The cost of running our MBQA model comprises two key components: (1) LLM API cost for training data collection and (2) GPU runtime expenses. 
For the first component, an upfront cost of approximately 31 USD is required to collect training data using \chkp{GPT-4o-mini} on the \data{MEDI2} dataset. This cost covers 1,000 text-question pairs for each of 10,000 questions, resulting in a total of 10 million text-question pairs. 
The cost scales down proportionally for fewer questions.
For the second component, we measure the training and inference time of our model on a single GTX 1080 Ti GPU. 
We estimate the cost based on a rental rate of 0.08 USD per hour.\footnote{Costs obtained from~\url{https://vast.ai/pricing/gpu/GTX-1080-TI}} 
Training for 3 million steps took around 36 hours, and inference times for the \data{MS MARCO} dataset varied by the number of dimensions: 48 hours for 2,000 dimensions, 63 hours for 4,000, 73 hours for 6,000, 79 hours for 8,000, and 90 hours for 10,000 dimensions.

\section{Implementation Details}
\label{app:experiment-details}

\begin{table}[t]
\centering
\caption{Hyperparameters used in our experiments.}
\label{tab:hyperparameters}
\vspace{-0.5em}
\begin{tabular}{lll}
  \toprule
  \textbf{Description} & \textbf{Symbol} & \textbf{Value} \\
  \midrule
  Encoding model & $Enc$ & \chkp{UAE-Large-V1} \\
  Generation model & $LLM$ & \chkp{GPT-4o-mini-2024-07-18} \\
  Number of clusters & $k$ & 5,000 \\
  Positive samples per cluster & $n_{p}$ & 6 \\
  Hard negative samples per cluster & $n_{h}$ & 18 \\
  Easy negative samples per cluster & $n_{e}$ & 18 \\
  Positive probe samples per question & $p_{p}$ & 5 \\
  Hard negative probe samples per question & $p_{h}$ & 3 \\
  Easy negatives probe samples per question & $p_{e}$ & 2 \\
  Deduplication threshold & $\theta$ & 0.8 \\
  Top questions per cluster & $t$ & 4 \\
  Learning rate of the MBQA Model & $\alpha$ & 1e-4 \\
  Binary classification threshold & $\tau$ & 0.5 \\
  \bottomrule
\end{tabular}
\end{table}

\subsection{Our CQG-MBQA Model}
\label{app:experiment-details:cqg-mbqa}

In the experiments, we train the proposed \algo{CQG-MBQA} framework using the \data{MEDI2} dataset \citep{muennighoff2024generative}. 
Detailed information about the model configuration and hyperparameters used in our framework is provided in Table \ref{tab:hyperparameters}.

\paragraph{Data Pre-processing}
We use the \data{MEDI2} dataset, downloaded from the HuggingFace repository at \chkp{GritLM/MEDI2}.\footnote{\url{https://huggingface.co/datasets/GritLM/MEDI2/tree/main}} 
We filter out files starting with \textit{task}, as they are unsuitable for the training corpus. 
From the remaining files, we extract both positive and negative instances of each data line. 
Since the \data{MEDI2} dataset contains instructions for queries and documents, we remove the instruction part, leaving only the content. We merge all positive and negative instances from the filtered corpus and run a simple exact deduplication to produce the final training text corpus.

\paragraph{Contrastive Question Generation}
The pre-processed training corpus contains 6.8 million unique texts. We encode these texts using the \algo{AnglE} encoding model (\chkp{UAE-Large-V1}) and normalize the embeddings. 
We then run \algo{KMeans} clustering \citep{kmeanspp} with $k=5,000$ clusters and default parameters, utilizing using the scikit-learn library,\footnote{\url{https://scikit-learn.org/stable/modules/generated/sklearn.cluster.KMeans.html}} accelerated by Intel(R) Extension for scikit-learn.\footnote{\url{https://github.com/intel/scikit-learn-intelex}} 
Once clustering is complete, we generate questions for each cluster according to the process described in Section \ref{sect:framework:generation}. 
We sample random positive examples from within the cluster, hard negatives from the three nearest clusters, and easy negatives from the remaining clusters. This process yields 9,614 deduplicated questions, which serve as the final embedding dimensions.

\paragraph{Multi-Task Binary Question Answering}
We train the MBQA model following the setup described in Appendix \ref{app:question-answering-performance}.

\begin{table}[t]
\centering
\caption{Model checkpoints used in our experiments.}
\label{tab:baselines}
\vspace{-0.5em}
\begin{tabular}{cc}
  \toprule
  \textbf{Model} & \textbf{Checkpoint}\\
  \midrule
\algo{BERT} & \chkp{bert-base-uncased} \\
\algo{GloVe} & \chkp{average\_word\_embeddings\_glove.6B.300d} \\
\algo{SimCSE (Unsup.)} & \chkp{unsup-simcse-bert-base-uncased} \\
\algo{SimCSE (Sup.)} & \chkp{sup-simcse-bert-base-uncased} \\
\algo{SBERT (New)} & \chkp{all-MiniLM-L12-v2} \\
\algo{OpenAI} & \chkp{text-embedding-ada-002} \\
\algo{AnglE} & \chkp{UAE-Large-V1} \\
\algo{BM25} & \chkp{bm25s} \\
  \bottomrule
\end{tabular}
\end{table}

\subsection{Baseline Models}
\label{app:experiment-details:baselines}


\paragraph{QAEmb-MBQA}
To ensure a fair comparison with \algo{QAEmb} \citep{Benara2024CraftingIE}, we modify our framework by replacing the CQG step with the example-based prompting method adopted from \algo{QAEmb}, while keeping all other parameters the same, except for setting the deduplication threshold $\theta=0.925$. 
Detailed prompts used for this version are provided in Appendix \ref{app:prompts:qaemb-question-generation}. 
This modification results in a total of 10,654 questions. 
The MBQA model is trained using the same approach as in our \algo{CQG-MBQA} framework.

\paragraph{Black-box Models}
For STS tasks, the results for \algo{SBERT (Ori.)} and \algo{USE} are sourced from \citep{reimers-gurevych-2019-sentence}, and the results for \algo{WhitenedCSE} are taken from the best-performing model in \citep{zhuo-etal-2023-whitenedcse}, all evaluated using the same metric. 
Table \ref{tab:baselines} shows the model checkpoints used for each black-box model. 
For STS, retrieval (excluding \data{MS MARCO} and \data{NewsSpectrum}), and clustering tasks, the results for \algo{BERT}, \algo{GloVe}, \algo{SimCSE (Unsup.)}, \algo{SimCSE (Sup.)}, \algo{SBERT (New)}, \algo{OpenAI}, and \algo{AnglE} are retrieved from the MTEB leaderboard.\footnote{\url{https://huggingface.co/spaces/mteb/leaderboard}} 
The retrieval results for \algo{BM25} (excluding the datasets \data{MS MARCO} and \data{NewsSpectrum}) are also retrieved from the MTEB leaderboard, while our own experiments were conducted for \data{MS MARCO} and \data{NewsSpectrum}.

\end{document}